# NEU-ESC: A Comprehensive Vietnamese dataset for Educational Sentiment analysis and topic Classification toward multitask learning

Phan Quoc Hung Mai[1*†], Quang Hung Nguyen[1*], Phuong Giang Duong[1], Hong Hanh Nguyen[1], Nguyen Tuan Long[1]

[1] College of Technology, National Economics University, Vietnam

[†]Corresponding author. E-mail: maphquochung@gmail.com

Contributing authors: {11222618,11221733,11222164}@st.neu.edu.vn; ntlong@neu.edu.vn

*These authors contributed equally to this work

**Abstract:** *In the field of education, understanding students' opinions through their comments is crucial, especially in the Vietnamese language, where resources remain limited. Existing educational datasets often lack domain relevance and student slang. To address these gaps, we introduce NEU-ESC, a new Vietnamese dataset for Educational Sentiment Classification and Topic Classification, curated from university forums, which offers more samples, richer class diversity, longer texts, and broader vocabulary. In addition, we explore multitask learning using encoder-only language models (BERT), in which we showed that it achieves performance up to 83.7% and 79.8% accuracy for sentiment and topic classification tasks. We also benchmark our dataset and model with other datasets and models, including Large Language Models, and discuss these benchmarks.*
*The dataset is publicly available at*: https://huggingface.co/datasets/hung20gg/NEU-ESC.

## 1. Introduction

Disruptions in the education sector, such as sudden policy changes, exam cancellations, or the shift to online learning during the COVID-19 pandemic, often trigger a surge in online public discourse [1, 2]. These events can amplify negative sentiment among students, parents, and educators, posing challenges to institutional credibility and public trust in the education system. In this context, social listening plays a crucial role in identifying and responding to the concerns of various stakeholders in the digital age [3, 4]. In education, particularly online learning environments, understanding students' desires through their comments is essential [5]. It helps monitor discussion forums, gain valuable insights from students' post comments, and avoid spam or unnecessary drama. This approach enables educators to tailor their teaching strategies and support systems more effectively based on real-time student sentiment.

Text classification and sentiment analysis are prominent tasks in Natural Language Processing (NLP) and have numerous applications across various domains. In education, it can be beneficial for analyzing and categorizing large volumes of student-generated content, such as forum posts and feedback. However, there is a notable lack of comprehensive datasets about education in Vietnam that offer deep classification of content, particularly from university forums. This gap presents a significant challenge for researchers and educators who aim to leverage NLP techniques to improve the educational experience in the Vietnamese context. A well-curated dataset with detailed content classification could provide valuable insights into student behaviors, preferences, and learning patterns specific to the Vietnamese educational system.

Inspired by these issues, our contributions are two-fold. Firstly, we propose a novel Vietnamese dataset called NEU-ESC for Educational Sentiment Analysis and Topic Classification, which offers more samples, more sentiment and topic classes, a larger sample length, and a larger vocabulary size than existing Vietnamese education datasets. Secondly, we provide a detailed analysis of the NEU-ESC dataset using a variety of Bidirectional Encoder Representations from Transformers (BERT) variants for both single-task and multitask classification schemes, which serve as initial benchmarks for our proposed dataset. Especially with multitask classification, we benchmark our proposed method and fine-tuned model with other works to demonstrate the effectiveness of our experiment.

In general, our research emphasizes two objectives:

- **Objective 1**: Creating the NEU-ESC dataset and evaluating the impact of incorporating the NEU-ESC dataset to address the current lack of generalization observed in the existing dataset.
- **Objective 2**: Experiment with different multitask learning techniques to BERT model and assess its superior performance in comparison to other models in the given tasks.

# 2. Related works

## 2.1. Language understanding

Understanding natural language, how humans communicate through speech and text, is a central challenge in artificial intelligence. NLP aims to bridge this gap by enabling machines to interpret, analyze, and generate human language. Among the most significant breakthroughs in this field is the development of Transformer architecture. Introduced by [6], the Transformer architecture quickly became the cornerstone of state-of-the-art NLP models. Their versatility extends beyond translation tasks, proving effective in tasks like summarizing long articles into concise pieces, analyzing the sentiment of online reviews, and answering complex questions posed in natural language. The transformer's success lies in its ability to capture intricate linguistic nuances, making it an indispensable tool for not only advancing current NLP applications but also unlocking new possibilities in natural language understanding and generation.

Bidirectional Encoder Representations from Transformers (BERT) [7] is a pretrained, unsupervised model trained using the English Wikipedia corpus. This bidirectionality sets it apart from earlier models that read text sequentially in one direction. BERT is trained using two unsupervised objectives: Masked Language Modeling (MLM), where random words are masked and predicted, and Next Sentence Prediction (NSP), which helps the model understand sentence-level relationships. Combined with transfer learning, BERT can be fine-tuned for a wide range of downstream tasks, such as question answering, sentiment analysis, and named entity recognition, without altering the model architecture. Its ability to generalize across tasks with minimal task-specific adaptation has made BERT a foundational model in modern NLP.

BERT has demonstrated remarkable effectiveness across a wide range of natural language processing tasks, especially in the domain of natural language understanding, enabling more accurate and context-aware language understanding in practical applications [8, 9]. In information retrieval and question answering systems, BERT enhances the ability to match user queries with relevant documents by capturing the nuanced meaning of language. [10] applied BERT to enhance document retrieval by integrating it with an information retrieval toolkit, enabling end-to-end neural search over large-scale document collections. BERT is experimented with [11] for multi-turn response selection in retrieval-based chatbots by introducing a speaker-aware extension (SA-BERT) that incorporates speaker change information and a disentanglement strategy to better model context in dialogue systems. Furthermore, BERT contributes to sentiment analysis, text classification, and named entity recognition tasks, providing valuable insights in domains such as social media monitoring, market research, and healthcare [12-14]. In educational settings, BERT has also been applied to tasks like automated grading [15, 16] and summarization of student feedback [17, 18]. The model's versatility and deep contextual understanding make it a foundational component in many state-of-the-art NLP systems.

Since its first release, there have been countless versions of BERT being developed, each of which might have slightly different structures and training methods, such as DistilBERT, RoBERTa, and MoEBERT [19-21], or using different types of datasets to create a BERT that is specific to a particular domain or language, such as Chinese BERT [22] a BERT model for Chinese languages, and Math-BERT.

In this research, we experimented with several common base versions of BERT variants in Vietnamese text classification to evaluate. XLM-Roberta (XLM-R), [23] developed by Facebook AI, is a multilingual model trained using a dataset weighing two terabytes from CommonCrawl data. It enhances the representation of languages with limited resources through up-sampling in training and vocabulary creation, resulting in a broader shared vocabulary. XLM-Roberta-base has 270 million parameters, and its larger version has 550 million parameters. PhoBERT [24] is the current state-of-the-art (SoTA) BERT for downstream Vietnamese tasks is phoBERT. PhoBERT is a pretrained model that shares a similar architecture with BERT and a training approach with RoBERTa. The base version of phoBERT has 135 million parameters, and it comes with two different versions: phoBERT-base and phoBERT-base-v2. VisoBERT [25] built on the powerful XLM-R architecture, it excels in pre-training on a rich and varied dataset of high-quality Vietnamese social media content. VisoBERT shows the best performance in five key natural language processing tasks: emotion recognition, hate speech detection, sentiment analysis, spam reviews detection, and hate speech spam detection. vELECTRA [26] is a fine-tuned version of the ELECTRA architecture [27], specifically adapted for the Vietnamese language using large-scale text corpora. It incorporates attentional recurrent neural networks during fine-tuning, resulting in improved performance on sequence labeling tasks like Part-of-Speech tagging and Named Entity Recognition.

## 2.2. Text classification and Sentiment analysis in Education

Text classification, a common task in NLP, involves assigning predefined labels or categories to text documents, sentences, or phrases based on their content. This task aims to automatically predict the class or category of the given text based on its context. Text classification plays an important role in educational management. Understanding the students' desires plays a crucial role in increasing the quality of education. With the help of AI, educators can now easily gather students' complaints from numerous posts and comments on different educational forums or platforms and make necessary changes in time. In the research [28], the authors used a so-called BERT-CNN-BiLSTM, along with the BERT-BiGRU of [29] and the combination of attention mechanism with CNN-GRU [30] to classify posts in the Stanford MOOC Posts dataset to decide whether the post in three groups A, B, and C is urgent or not. [31] used data mining techniques to understand campus trends, which highlights the events that attract considerable attention, thus proving beneficial for both institutional leadership and technology providers aiming to implement smart campus initiatives.

In Vietnam, there have been several major contributions to this field of study [32, 33], all of which focus on analyzing comments on social media. Datasets in these studies act as a benchmark for NLP tasks in Vietnam and various research [32-34] have attempted to maximize the accuracy score by testing them. [32] focuses on analyzing hate speech and comments on social media, while [35] proposes two different datasets for sentiment analysis, namely, VLSP 2016 for news and VLSP 2018 for hotels and restaurants. The UIT-VSFC dataset of [36] comprises over 16,000 student feedback entries, offering two different label schemes for sentiment-based and topic classification tasks. This dataset is valuable for educational research, providing insights into student experiences and perceptions, and is mainly used as a benchmark for text classification. However, none of these datasets fully meet the requirements for analyzing open education forums on social media platforms. Particularly, while the UIT-VSFC dataset is relatively formal and lacks common slang, the others are not closely related to the education sector. However, none of the datasets match the requirements for an open education forum in social media platforms, as while VSFC is relatively more formal and lacks common slang, the others are not closely related to education.

Recently, Large Language Models (LLMs), which are known for their advanced capabilities in understanding and generating text, have become a standard for evaluating linguistic tasks [37]. To benchmark their performance, any work related to language understanding should conduct at least a brief comparison between the proposed method and popular LLMs to evaluate their ability to perform specific tasks. Benchmarking against LLMs is essential because these models set a high baseline for multilingual understanding, providing a clear reference point for assessing the effectiveness of our method.

## 2.3. Techniques in fine-tuning BERT models

There are numerous approaches using transfer learning for text-classification tasks in general, with BERT, and among them, the two most common techniques are using solely a feed-forward layer (BERT-Linear) for classification, with the support of the Huggingface library [38-41], or using one or multiple CNN layers before a classification layer (BERT-CNN) [42, 43]. These types of networks will be connected after the last hidden layer of BERT, as demonstrated by following the diagrams of **Figure 1**.

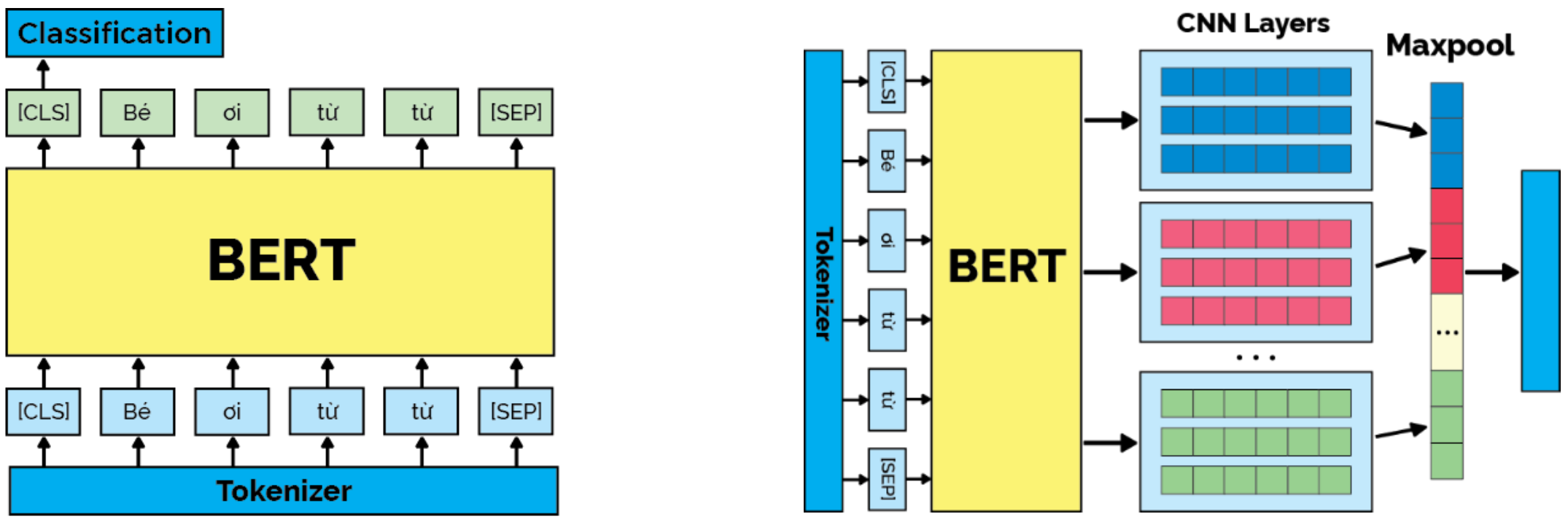

*a. BERT-Linear* *b. BERT-CNN*

***Figure 1.** Common fine-tuning BERT architectures.*

In the BERT-Linear approach **(Figure 1a)**, the output of token <CLS> is processed through a feed-forward block to generate the predicted label. In the BERT-CNN approach **(Figure 1b)**, the embedded vectors from BERT traverse a CNN block comprising multiple layers with varying filter sizes, in which CNN layers employ a max-pooling function to extract key information from each filter. These features are then linearized and concatenated, and the combined result passes through a feed-forward layer to produce the final outputs.

Rather than only fine-tuning on an individual task, one can alternatively make use of multitask learning to update BERT. Since it is useful for multiple and related tasks to be learned jointly at the same time, so that the model can share knowledge between different tasks, [44] introduced a model that incorporates a framework combining a shared BERT layer with task-specific prediction heads. Therefore, the use of a shared BERT layer enables the model to leverage cross-task data and generate more generalized representations. Meanwhile, [45] attempted to perform multitask learning by adding each loss together on the tasks of category classification and named entity recognition:

$$\mathcal{L}_{total} = \sum_{i} \alpha_i \mathcal{L}_{task_i}$$

As of the paper [46], the authors proposed the round-robin strategy for training on the total loss function, where the optimizer steps once per task and rotates between tasks on batch levels. Similarly, [44] used the same strategy to loop over each batch of different tasks to update the parameters of the models, and this approach performed well in improving the accuracy rate of each task.

Aggressive fine-tuning frequently leads to overfitting, resulting in the model's failure to generalize effectively to unseen data. To address this issue in a principled manner, [47] proposes a technique: Smoothness-inducing regularization (SMART loss regularization), which would effectively manage the model's complexity.

Specifically, in the context of the model $f(\cdot, \theta)$ and $n$ data points from the target task, represented as $\{(x_i, y_i)\}_{i=1}^{n}$ where $x_i$ signifies the embedding of input sentences derived from the initial embedding layer of the language model (BERT), and yi denotes the corresponding labels, [47] the approach essentially addresses the following optimization problem for fine-tuning:

$$min_{\theta} = \mathcal{L}(\theta) + \lambda_s \mathcal{R}_s(\theta)$$

Where $\mathcal{L}(\theta)$ is the loss function defined by calculating the sum of the target tasks' loss $\ell(\cdot,\cdot)$: $\mathcal{L}(\theta) = \frac{1}{n}\sum_{i=1}^{n} \ell(f(\cdot,\theta), y_i)$. With $\lambda_s > 0$ is a tuning parameter and $\mathcal{R}_s(\theta)$ is the smoothness-inducing regularizer defined as:

$$\mathcal{R}_s(\theta) = \frac{1}{n}\sum_{i}^{n} \max_{\|\tilde{x}_i - x_i\|_{p<\epsilon}} \ell(f(\tilde{x}_i,\theta), f(x_i,\theta))$$

Where $\epsilon > 0$ is a tuning parameter. And that for classification tasks, $f(\cdot,\theta)$ outputs a probability simplex and $\ell_s$ is chosen as a symmetrized KL-divergence. [47] also proposed a class of Bergman Proximal Point Optimization methods to minimize the above function. These methods employ a strong regularizer at each update, which discourages the model from drastic parameter adjustments.

## 3. Dataset

In this section, we describe our dataset collection, cleansing, and general statistical analysis of the composed dataset.

### 3.1. Data collection and curation

The raw data for our study was collected from a variety of online sources, with a primary focus on university-related Facebook groups and community forums where students frequently engage in discussions. These platforms were chosen due to their high activity levels and the presence of organic, user-generated content reflecting real student opinions and experiences. To capture content with potentially harmful or offensive language, particularly hate speech, we additionally scraped comments from less moderated forums such as XamVN and Voz, which are known for their more unfiltered and informal discourse.

As anticipated, the linguistic quality of the raw data presented several challenges. The use of teen code (slang, abbreviations, and phonetic spellings), as well as a significant number of typographical errors, was pervasive. This is characteristic of informal online communication, especially among younger demographics. Consequently, a comprehensive preprocessing pipeline was essential. This pipeline involved cleaning the text by removing redundant characters, emojis, special symbols, and other forms of noise that could interfere with downstream natural language processing tasks.

During this preprocessing phase, we also identified a substantial use of acronyms and shorthand expressions, particularly related to academic subjects and majors, either in Vietnamese or English. These abbreviations are widely used by students for convenience in casual conversation. For instance, 'qtkd' is a common shorthand for *Quản trị kinh doanh* (Business Administration), while 'dsa' refers to *Data Structure and Algorithm.* To preserve the semantic meaning of such terms and enhance the consistency of our records, we manually constructed a mapping dictionary that translates these acronyms into their full textual representations.

After applying the full data curation and preprocessing pipeline, we compiled a final dataset, referred to as the NEU dataset for Educational Sentiment analysis and topic Classification (NEU-ESC), consisting of nearly 33,000 samples. This dataset provides a rich and realistic resource for tasks such as sentiment analysis, hate speech detection, and opinion mining within the context of the Vietnamese higher education system.

## 3.2. Dataset composition

***Table 1.*** *Sentiment labels statistics of NEU-ESC*

| *Label* | *Sentiment* | *Count* | *Percentage* | *Mean Length* |
|---|---|---|---|---|
| *0* | *Neutral* | *22,773* | *69.08%* | *23.21* |
| *1* | *Positive* | *4,148* | *12.58%* | *24.29* |
| *2* | *Negative* | *5,250* | *15.77%* | *34.30* |
| *3* | *Toxic* | *845* | *2.56%* | *22.78* |

On the sentiment analysis task (**Table 1**), users' comments will belong to each of these four categories: Neutral, Positive, Negative, and Toxic. If a comment is Neutral, it rarely expresses the commenter's emotions on a specific topic. If a comment is Positive or Negative, then it either shows students' satisfaction or dissatisfaction with the problems discussed. Lastly, there is a toxic label for comments with strong, hateful words or the use of offensive language aimed at others. For the topic classification task, students' comments will be categorized into nine different groups, showing their relevance to the chosen topic: Spam, News, Academic, Neutral, Service, Job & Recruitment, Personal Affairs, Social Affairs, Helping & Sharing, and Club & Events.

The dataset is predominantly composed of neutral comments, which make up 69.08% of the data, with an average length of 23.21 words. Positive comments account for 12.58%, slightly longer on average at 24.29 words. Negative comments represent 15.77% but stand out with the highest average length of 34.30 words, suggesting users tend to express negative opinions in more detail. Toxic comments are the least frequent, comprising only 2.56% of the data, with an average length similar to neutral comments at 22.78 words. This distribution indicates a largely neutral tone in educational discussions, with a relatively small but notable portion of negative and toxic sentiment.

***Table 2.*** *Topic classification labels statistics of NEU-ESC*

| Label | Classification | Count | Percentage | Mean Length |
|---|---|---|---|---|
| 0 | Spam | 405 | 1.23% | 22.71 |
| 1 | News | 902 | 2.74% | 59.55 |
| 2 | Academic | 10,512 | 31.89% | 29.62 |
| 3 | Other | 14,402 | 43.69% | 11.40 |
| 4 | Service | 2,358 | 7.15% | 30.94 |
| 5 | Jobs & Recruitment | 808 | 2.45% | 55.14 |
| 6 | Personal Affairs | 1,478 | 4.48% | 33.17 |
| 7 | Social Affairs | 769 | 2.33% | 67.11 |
| 8 | Help & Share | 670 | 2.03% | 37.03 |
| 9 | Club & Events | 662 | 2.01% | 68.82 |

For topic classification (**Table 2**), the dataset is categorized into ten content-based labels reflecting the variety of discussions found in online educational communities. These labels include Spam, News, Academic, Other, Service, Jobs & Recruitment, Personal Affairs, Social Affairs, Help & Share, and Club & Events. Each label captures a distinct communication intent from academic inquiries and student services to personal posts and social interactions, allowing for a more nuanced understanding of user-generated content in the education domain.

From a statistical perspective, the ‘Other’ category dominates the dataset, making up 43.69% of the entries, though it has the shortest average message length (11.40 words), suggesting many of these posts may be brief or contextually ambiguous. Academic content follows at 31.89%, with a moderate average length of 29.62 words. In contrast, categories like Club & Events (68.82 words) and Social Affairs (67.11 words) have the longest average message lengths, indicating more detailed or descriptive posts. Less common categories, such as Spam (1.23%) and Help & Share (2.03%), still add important diversity to the dataset, capturing both irrelevant and collaborative content. This distribution highlights both the breadth and depth of communication within educational online spaces.

***Table 3.*** *NEU-ESC dataset split*

| **Subset** | **Number of Samples** | **Actual Ratio** | **Mean Length** |
|---|---|---|---|
| Training set | 23,048 | 0.699 | 24.81 |
| Test set | 6,613 | 0.200 | 26.02 |
| Validation set | 3,305 | 0.101 | 25.12 |

NEU-ESC contains a total of 32,966 samples. The distribution of the number of each sentiment and topic label pair is divided into train, validation, and test sets with a ratio of 7:1:2, corresponding to 23,048 for the training set, 3,305 for the validation set, and 6,613 for the test set (see **Table 3**).

## 3.3. Datasets benchmarks

In addition to our own dataset, we also utilize several widely used benchmark datasets in the same field to evaluate our models.

Vietnamese Students’ Feedback Corpus (UIT-VSFC) [36] is a dataset collected from over 16,000 students’ feedback. The dataset comprises three fields: comment text, sentiment analysis, and topic classification. Sentiment analysis categorizes data as Positive, Negative, or Neutral, while topic classification identifies themes such as training programs, lecturers, faculty, and other relevant subjects. Vietnamese Hate Speech Detection (ViHSD) [33] contains 33.400 annotated comments on social media platforms. The dataset classifies comments as clean (non-hate), offensive, or hate. Vietnamese Constructive and Toxic Speech Detection (ViCTSD) [32] is a dataset for constructive and toxic speech detection in Vietnamese. It consists of 10,000 human-annotated comments. The dataset aims to classify 4 labels for toxic level and 2 labels to identify whether the comment is constructive or non-constructive.

***Table 4.*** *Comparative statistics of NEU-ESC with other datasets*

| Criteria | UIT-VSFC | ViHSD | ViCTSD | NEU-ESC* |
|---|---|---|---|---|
| Total samples | 16,175 | 33,400 | 10,000 | 32,966 |
| Number of tasks | 2 | 1 | 2 | 2 |
| Number of labels | 7 | 3 | 4 | 14 |
| Avg. words/sample | 14.23 | 11.40 | 29.74 | 25.09 |
| Vocabulary size | 2,877 | 12,183 | 7,785 | 12,759 |

The comparative statistics presented in **Table 4** highlight the strengths and distinctiveness of the NEU-ESC dataset as a comprehensive benchmark for Vietnamese NLP tasks. Despite having a sample size comparable to ViHSD, NEU-ESC contains more diverse vocabulary (12,759 unique tokens), suggesting greater lexical coverage and domain variety. Additionally, the dataset maintains a high average sentence length (25.09 words per sample), second only to ViCTSD, indicating a higher level of linguistic complexity. These features make NEU-ESC a premier benchmark for training and evaluating robust models in Vietnamese sentiment and topic classification, particularly for applications involving long-form, user-generated content in educational and social contexts.

Our dataset significantly expands upon the UIT-VSFC dataset [36], which is limited to three sentiment labels and only four topic categories. In contrast, the proposed NEU-ESC dataset provides a richer annotation schema with more diverse sentiment and topic labels, enabling finer-grained analysis of student-generated content. Furthermore, while UIT-VSFC is confined to student feedback, NEU-ESC covers a much broader range of discussion contexts within online educational platforms. Another distinguishing characteristic of NEU-ESC is its inclusion of a greater number of lengthy comments, particularly those exceeding 50 words, thereby offering more complex linguistic structures for analysis. **Table 5** presents a detailed statistical comparison between NEU-ESC and UIT-VSFC, highlighting the expanded scope and improved representativeness of our corpus.

***Table 5.*** *Benchmark NEU-ESC with UIT-VSFC*

| Criteria | UIT-VSFC | NEU-ESC* |
|---|---|---|
| Total samples | 16,175 | 32,966 |
| Sentiment labels | 3 | 4 |
| Topic labels | 4 | 10 |
| Avg. words/sample | 14.23 | 25.09 |
| 1-10 words | 7,232 | 13,638 |
| 10-20 words | 6,061 | 9,934 |
| 20-50 words | 2,702 | 6,169 |
| 50+ words | 180 | 3,225 |
| Vocabulary size | 2,877 | 12,759 |

## 4. Experiment

With the main goal to maximize the performance of models on two main tasks on our dataset: Topic classification and Sentiment analysis, we conducted different experiments to evaluate the accuracy of different approaches. In this section, we detail our implementation of BERT-base models for single tasks and of the various paths considered to leverage the performance of multitask learning.

### 4.1. Approach

For the experiments, we first implemented single-task fine-tuning with BERT and BERT-CNN on three BERT models: BERT-base [7], XLM-Roberta-base [23], phoBERT-base-v2 [48], vELCTRA [26], and VisoBERT [25] for the two main tasks. This allows us to assess the performance of different BERT models on different tasks and then pick the best models for further fine-tuning in multitask learning. Also, by learning each task individually, we can indicate the task-specific benefits and drawbacks of fine-tuning BERT models across multiple datasets simultaneously.

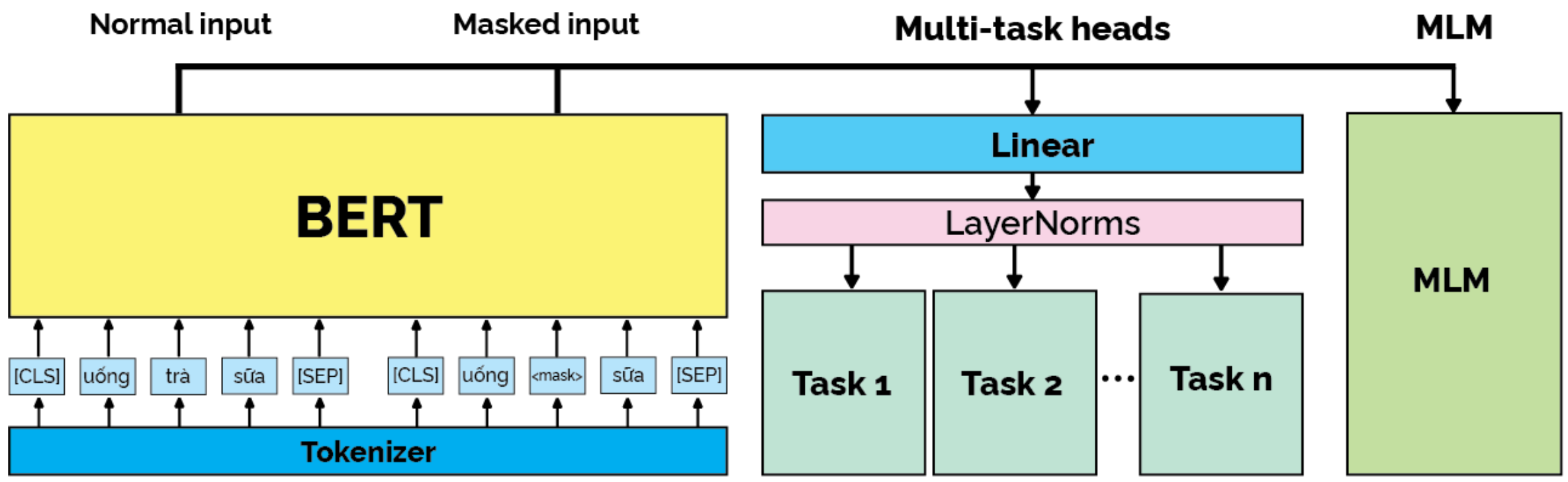

***Figure 2.*** *Illustration of a multitask model.*

Our Multitask BERT-base model (**Figure 2**) adopts an approach where a given tokenized sentence undergoes a BERT layer, followed by a linear layer, and then adds layer norms. After this process, the output is directed to different layers (so-called heads) for different tasks. Specifically, in our work, the output vectors will be passed through two distinct fully connected layers (two heads), each tailored for specific tasks: topic classification and sentiment analysis. This novel architecture enables the model to simultaneously learn and classify information for both tasks using a unified sentence representation.

To further enhance performance, we incorporate a masked language predicting head into the model (MLM), as extended to the idea of [7, 25, 49], contributing to improved language understanding and task-specific capabilities. Specifically, the MLM head is set up as follows: 30% of the tokens are chosen to be masked randomly, in which 80% of these tokens are labeled as <MASK>, 10% are mislabeled, and the remaining 10% are left unchanged. The task of this head is to predict words that have been replaced by <MASK> and recognize tokens that have been mislabeled.

In our experiments, we evaluated the model both with and without the masked language predicting head for comprehensive analysis. In general, we experimented with multitask learning with the 2-head and 3-head BERT-base model, where the 2-head consists of classifying two main tasks, and the 3-head model consists of an additional masked language model predicting the head. This combination of BERT and masked language prediction collectively enhances the multitask capabilities of the model, making it adept at handling diverse natural language processing tasks.

The loss function of the whole multitask model is calculated by adding all losses of each head together as follows:

$$\mathcal{L}_{total} = \mathcal{L}_{classification} + \mathcal{L}_{sentiment} + \frac{\mathcal{L}_{MLM}}{\sigma}$$

Where $\sigma$ is the average token per sentence. We observe that the value of $\mathcal{L}_{MLM}$ tends to dominate the overall loss, being significantly larger than both $\mathcal{L}_{classification}$ and $\mathcal{L}_{sentiment}$. This imbalance causes the model to prioritize token prediction over the primary tasks. To mitigate this effect, since the MLM head primarily serves as a regularization buffer during training, we reduce its impact by scaling its loss down, dividing it by the average number of masked tokens per sentence.

In addition, we investigate the effect of incorporating SMART loss [47] during the fine-tuning process of the language model. Specifically, this regularization encourages the model to produce stable predictions under small, adversarial perturbations to the input, thereby enhancing its robustness. For each task **i**, the modified loss function is defined as:

$$\mathcal{L}_i = \mathcal{L}_i(\theta) + \lambda_s \mathcal{R}_s(\theta)$$

Where $\mathcal{L}_i(\theta)$ is the original task loss, $\mathcal{R}_s(\theta)$ is the SMART regularization term, and $\lambda_s$ is a hyperparameter that controls the strength of the regularization. This formulation is based on the assumption from the original SMART paper that smoothness in the model's output space helps mitigate overfitting, particularly when fine-tuning large pre-trained language models on limited labeled data.

Using our Evaluating set, we also experimented on the common LLMs: GPT 4o [50] and Claude 4 Sonnet [51] as a benchmark for BERT models. To comprehensively assess the performance of these models, we conducted experiments under both zero-shot and few-shot settings. In the zero-shot setting, the models received no prior examples and were expected to generalize directly from the prompt. In contrast, the few-shot setting involved providing the models with labeled examples to guide their responses. In this experiment, we set 10 examples from the training dataset that cover all topics and sentiment labels. This approach allowed us to evaluate each model's ability to understand and generalize across different levels of contextual information.

In total, we conducted four experiments to compare different fine-tuning strategies for BERT on two main tasks: topic classification and sentiment analysis. The setups included single-task learning, two-task multitask learning, two-task multitask learning with SMART loss, and multitask learning with an auxiliary masked language modeling (MLM) objective.

## 4.2. Result

In this section, we report the results of fine-tuning various BERT-based models on the NEU-ESC dataset and provide a benchmark with Large Language Models (LLMs). All experimental results are presented in **Table 6**, which illustrates the performance of different model architectures on both the sentiment analysis and topic classification tasks.

Firstly, we evaluate each model fine-tuned on a single task and then select the best-performing models for further evaluation in a multitask fine-tuning setting. As shown in the results, VisoBERT excels in the topic classification task, while phoBERT$_{base\text{-}v2}$ performs best in sentiment analysis among the single-task models. Additionally, XLM-RoBERTa$_{base}$ demonstrates its potential by achieving competitive performance, whereas BERTbase and vELECTRA perform relatively poorly. For subsequent experiments, we proceed with multitask fine-tuning using phoBERT$_{base\text{-}v2}$, XLM-RoBERTa$_{base}$, and VisoBERT to explore potential performance improvements.

For the 2-task models, a slight improvement was observed. Specifically, phoBERT$_{base\text{-}v2}$ achieved the best performance across both tasks in this setting. Compared to its single-task counterpart, phoBERT$_{base\text{-}v2}$ 's accuracy increased by 0.1% in the sentiment analysis task and by 0.6% in the topic classification task. These results highlight the potential benefits of applying multitask learning to enhance model performance.

When fine-tuning the multitask models using SMART loss, we observe a significant improvement in performance across both tasks. Notably, VisoBERT stands out in this setting, achieving accuracies of 83.58% for sentiment analysis and 79.80% for topic classification—its highest performance in all configurations. One can observe that all models have increments in their performance. For this setting, VisoBERT achieves the best performance at 79.80%, 63.13%, and 80.55%, respectively, for accuracy, mF1, and wF1 on the topic classification task among all experimented models.

When combining the two main tasks with masked language modeling (MLM), the improvements over single-task models were not significant. Only slight and inconsistent performance gains were observed. However, when multitask learning was combined with both MLM and SMART loss, the improvements became more pronounced. Notably, phoBERT$_{base-v2}$ achieved an accuracy of 83.66% on the sentiment analysis task—the highest among all models tested. This indicates that integrating SMART loss with MLM can further enhance the performance of BERT-based models in multitask settings. Nevertheless, it is important to note that incorporating MLM substantially increases computational complexity. This is because it requires two forward and backward passes through the BERT model, along with additional computations for next-token prediction, making the overall training process significantly more resource-intensive. These results clearly demonstrate the effectiveness of SMART loss in enhancing the learning capability of BERT-based models. The consistent performance gains across different architectures suggest that incorporating SMART loss can lead to more robust and generalizable models in multitask learning scenarios.

Despite LLMs' general superiority, they failed to outperform the BERT-based model in our evaluation. This holds in both zero-shot and few-shot settings. While LLMs may achieve relatively high accuracy in sentiment classification, their performance still falls short compared to the multi-task PhoBERT model. However, the accuracy of the highest LLM only reaches 77.84% for sentiment analysis and 42.31% for sentiment analysis, which is far apart from any BERT model. It is important to note that sentiment classification involves only 4 labels, whereas topic classification includes a more complex label space with 10 distinct classes. This greater label diversity partly explains why LLMs performed poorly in the topic classification task.

*__Table 6.__ Result of different models on the NEU-ESC dataset (%)*

| **Model type** | **Model** | **Sentiment Analysis** | | | **Topic Classification** | | |
|---|---|---|---|---|---|---|---|
| | | Accuracy | mF1 | wF1 | Accuracy | mF1 | wF1 |
| **Single task** | $BERT_{base}$ | 77.34 | 70.01 | 76.68 | 75.32 | 57.65 | 75.06 |
| | vELECTRA | 78.47 | 71.73 | 78.12 | 75.40 | 53.46 | 74.39 |
| | $XLM_R_{base}$ | 81.46 | 75.91 | 81.27 | 77.58 | 60.87 | 77.60 |
| | VisoBERT | **82.78** | 75.79 | **82.17** | 77.94 | 60.65 | 78.06 |
| | $phoBERT_{base\text{-}v2}$ | 81.75 | **77.70*** | 82.02 | **78.65** | **62.73** | **78.36** |
| **2-task** | $XLM_R_{base}$ | 82.19 | 75.50 | 81.39 | 77.36 | 60.62 | 77.89 |
| | VisoBERT | 82.63 | 76.17 | 82.58 | 78.56 | 61.66 | 78.68 |
| | $phoBERT_{base\text{-}v2}$ | **82.88** | **75.99** | **82.85** | **79.25** | **62.11** | **79.59** |
| **2-task & SMART** | $XLM_R_{base}$ | 83.03 | 76.74 | 83.69 | 79.54 | 62.90 | 80.17 |
| | VisoBERT | **83.58** | 77.17 | **84.19*** | **79.80*** | **63.13*** | **80.55*** |
| | $phoBERT_{base\text{-}v2}$ | 83.26 | **77.52** | 83.66 | 79.31 | 62.79 | 80.04 |
| **2-task & MLM** | $XLM_R_{base}$ | 82.49 | 76.38 | **82.94** | 77.86 | 60.33 | 78.29 |
| | VisoBERT | 82.23 | 75.15 | 82.70 | 78.38 | 60.79 | **78.87** |
| | $phoBERT_{base\text{-}v2}$ | **82.67** | **77.03** | 82.72 | **78.48** | **62.33** | 78.44 |
| **2-task & MLM & SMART** | $XLM_R_{base}$ | 82.65 | 75.97 | 83.51 | 78.96 | 60.93 | 79.85 |
| | VisoBERT | 83.12 | 76.74 | 83.57 | 79.25 | 62.19 | 79.88 |
| | $phoBERT_{base\text{-}v2}$ | **83.66*** | **77.34** | **84.15** | **79.49** | **62.58** | **80.04** |
| **LLM** | GPT 4o zero-shot | 72.81 | 49.70 | 72.83 | 14.52 | 22.05 | 11.93 |
| | Claude 4 zero-shot | 74.32 | 66.05 | 74.06 | 39.68 | 29.65 | 42.61 |
| | GPT 4o few-shot | 76.05 | 63.93 | 74.79 | 17.72 | 26.21 | 16.94 |
| | Claude 4 few-shot | **77.84** | **70.89** | **75.95** | **42.31** | **32.73** | **44.83** |

For error analysis, **Figure 3** shows the confusion matrix for one selected BERT model: 2-task & SMART VisoBERT. For sentiment analysis, the model performs best at identifying *Neutral* samples, but frequently misclassifies *Positive* and *Negative* sentiments as *Neutral*, suggesting a bias toward the majority class. *Toxic* sentiment is poorly predicted, with most samples misclassified, likely due to class imbalance. In the topic classification task, the model shows strong performance on *Other* and *Academic* classes, with relatively high true positive rates, but struggles with less frequent categories like *Jobs & Recruitment*, *Help & Share*, and *Club & Events*, often confusing them with more dominant classes such as *Other* or *Service*. This indicates the model captures frequent patterns well but needs improvement on underrepresented or semantically overlapping classes.

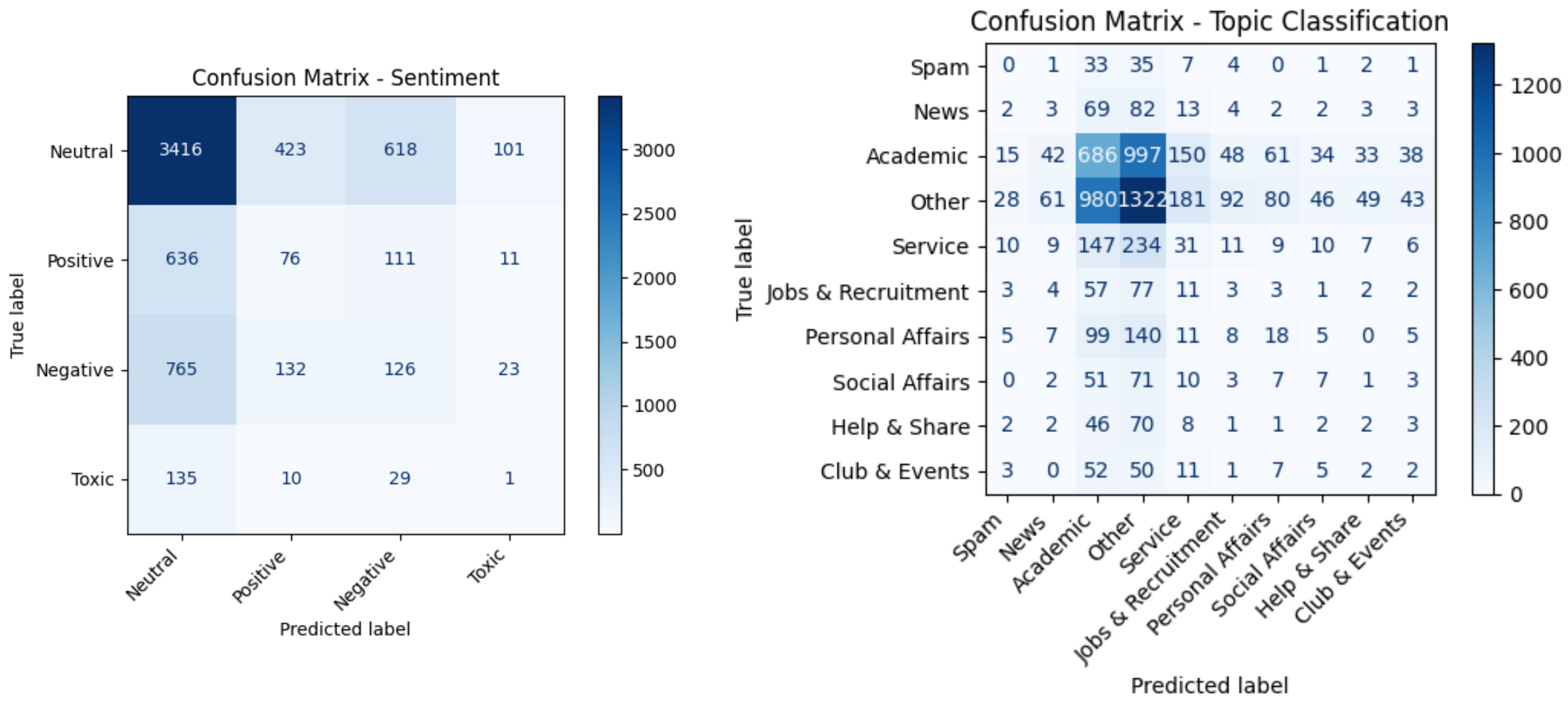

***Figure 3.*** *Confusion matrix of 2-task & SMART VisoBERT*

## 5. Discussion and Conclusion

### 5.1. Discussion

Given Vietnam's current lack of a large and comprehensive dataset that fully expresses students' thoughts and behaviors, our proposed dataset can contribute by being an open-source project soon. Even though data concerning students' behaviors can be collected through various social media platforms and by a wide range of services, Facebook data offers the most accurate representation of their desires. This is because Facebook is currently the most widely used social media platform in Vietnam [56], with most of its users belonging to the younger segment of the population. To ensure data quality, our dataset will be frequently updated to capture the latest university trends and adapt to schools' education. After being published, our dataset might be put in use to train other similar models or used for other classification tasks. Not to mention a community of business and educational analysts utilizing our datasets to adapt their policies accordingly to students' needs. However, the greatest contribution will come in the form of further research to enhance this university life monitoring system, as the data is cleaned up to standards and preprocessed carefully for model training. From the data collected, more insights from students' posts or comments could be drawn, giving researchers valuable ideas and, as a result, preventing unnecessary drama on social media.

Despite its benefits, the adoption of multi-task models has not been widespread, particularly among Vietnamese researchers. Through the integration of common Neural Network architectures on transfer learning, we have managed to achieve remarkable results. As we deploy increasingly sophisticated networks, we anticipate that multi-task models will exhibit significantly improved performance beyond our current capabilities.

Our initial analysis further reveals that LLMs often confuse similar labels and struggle to capture the nuanced or implicit meanings within sentences. This presents a dilemma: improving LLM performance typically requires more elaborate and detailed prompts, which can increase inference costs. Alternatively, fine-tuning smaller language models (e.g., Qwen [57], LLaMA [58]) might boost task-specific performance, but comes at the expense of higher computational demands. However, to avoid getting out of scope, the optimal use of LLMs in such a downstream task would need deeper research. Therefore, for certain downstream tasks, especially those involving structured label spaces or subtle linguistic distinctions, BERT-like models continue to demonstrate strong efficiency and reliability.

## 5.2. Conclusion and Future Work

In conclusion, the research focuses on creating a new dataset from NEU students on social media platforms, addressing common drawbacks of existing datasets. The proposed dataset, capturing students' thoughts and behaviors on Facebook, is positioned as a valuable resource for understanding and adapting to university trends, with potential applications in training other models and informing policies for businesses and educational institutions. The study employs a multitask approach, fine-tuning language models, with a 3-head model based on phoBERT and reinforcement of SMART loss regularization, emerging as the best performer. The findings indicate that this model excels in sentiment analysis and topic classification, outperforming other configurations and even surpassing current state-of-the-art models in VSFC and ViHSD datasets.

Despite the superiority of this dataset over others, the dataset is currently imbalanced. Most comments focused on topics related to other or academic within the confines of college life. This limited scope presents a significant challenge for developing broadly applicable models. In the future, to address this issue, it is crucial to expand the dataset to encompass a wider range of subjects and perspectives. A more diverse dataset would ideally include comments from various age groups, professional backgrounds, and life experiences beyond the academic sphere. This could involve gathering input on topics such as career development, personal finance, healthcare, technology, social issues, and global events. By broadening the scope of the dataset, we can capture a more accurate representation of real-world conversations that we consider as our future work.

Furthermore, other training techniques can also be applied for training multitask BERT. For example, dealing with the update of layers for two or more conflicting gradients, PCGrad [59] is a great tool for resolving the conflict. However, the magnificence of PCGrad comes with an extreme $O(n^2)$ computational requirement, therefore, in the scope of this paper, we are not yet experimenting with the advances of this technique due to a lack of resources. Also, there are numerous fine-tuning approaches to try out within multitask learning, such as the expansion of SMART loss: ALICE and ALICE++, or applying Adapter Modules [60, 61].

## References

1. Karalis, T., *Planning and evaluation during educational disruption: Lessons learned from COVID-19 pandemic for treatment of emergencies in education.* European Journal of Education Studies, 2020.
2. Giroux, H.A., *Higher education and the politics of disruption.* Chowanna, 2020(54 (1)): p. 1-20.
3. Giachanou, A. and F. Crestani, *Like it or not: A survey of twitter sentiment analysis methods.* ACM Computing Surveys (CSUR), 2016. **49**(2): p. 1-41.
4. Stewart, M.C. and C.L. Arnold, *Defining social listening: Recognizing an emerging dimension of listening.* International journal of listening, 2018. **32**(2): p. 85-100.
5. Shaik, T., et al., *Sentiment analysis and opinion mining on educational data: A survey.* Natural Language Processing Journal, 2023. **2**: p. 100003.

6. Vaswani, A., et al., *Attention is all you need.* Advances in neural information processing systems, 2017. **30**.
7. Devlin, J., et al., *BERT: Pre-training of Deep Bidirectional Transformers for Language Understanding.* 2019 Conference of the North {A}merican Chapter of the Association for Computational Linguistics: Human Language Technologies, 2019. **1**: p. 4171-4186.
8. Hambarde, K.A. and H. Proenca, *Information retrieval: recent advances and beyond.* IEEE Access, 2023. **11**: p. 76581-76604.
9. Soleimani, A., C. Monz, and M. Worring. *Bert for evidence retrieval and claim verification*. in *European Conference on Information Retrieval*. 2020. Springer.
10. Yilmaz, Z.A., et al. *Applying BERT to document retrieval with birch*. in *Proceedings of the 2019 Conference on Empirical Methods in Natural Language Processing and the 9th International Joint Conference on Natural Language Processing (EMNLP-IJCNLP): System Demonstrations*. 2019.
11. Gu, J.-C., et al. *Speaker-aware BERT for multi-turn response selection in retrieval-based chatbots*. in *Proceedings of the 29th ACM International Conference on Information & Knowledge Management*. 2020.
12. Khare, Y., et al. *Mmbert: Multimodal bert pretraining for improved medical vqa*. in *2021 IEEE 18th international symposium on biomedical imaging (ISBI)*. 2021. IEEE.
13. Sousa, M.G., et al. *BERT for stock market sentiment analysis*. in *2019 IEEE 31st international conference on tools with artificial intelligence (ICTAI)*. 2019. IEEE.
14. Shaik Vadla, M.K., M.A. Suresh, and V.K. Viswanathan, *Enhancing product design through AI-driven sentiment analysis of Amazon reviews using BERT.* Algorithms, 2024. **17**(2): p. 59.
15. Sharma, A. and D.B. Jayagopi, *Modeling essay grading with pre-trained BERT features.* Applied Intelligence, 2024. **54**(6): p. 4979-4993.
16. Zhu, X., H. Wu, and L. Zhang, *Automatic short-answer grading via BERT-based deep neural networks.* IEEE Transactions on Learning Technologies, 2022. **15**(3): p. 364-375.
17. Masala, M., et al. *Extracting and clustering main ideas from student feedback using language models*. in *International Conference on Artificial Intelligence in Education*. 2021. Springer.
18. Morris, W., et al., *Formative feedback on student-authored summaries in intelligent textbooks using large language models.* International Journal of Artificial Intelligence in Education, 2024: p. 1-22.
19. Liu, Y., et al., *Roberta: A robustly optimized bert pretraining approach.* arXiv preprint arXiv:1907.11692, 2019.
20. Sanh, V., et al., *DistilBERT, a distilled version of BERT: smaller, faster, cheaper and lighter.* arXiv preprint arXiv:1910.01108, 2019.
21. Zuo, S., et al., *Moebert: from bert to mixture-of-experts via importance-guided adaptation.* 2022 Conference of the North American Chapter of the Association for Computational Linguistics: Human Language Technologies, 2022: p. 1610–1623.
22. Cui, Y., et al., *Pre-training with whole word masking for chinese bert.* IEEE/ACM Transactions on Audio, Speech, and Language Processing, 2021. **29**: p. 3504-3514.
23. Conneau, A., et al., *Unsupervised cross-lingual representation learning at scale.* 58th Annual Meeting of the Association for Computational Linguistics, 2020: p. 8440–8451.
24. Nguyen, D.Q. and A.T. Nguyen, *PhoBERT: Pre-trained language models for Vietnamese.* Findings of the Association for Computational Linguistics: EMNLP 2020, 2020: p. 1037–1042.
25. Nguyen, Q.-N., et al., *ViSoBERT: A Pre-Trained Language Model for Vietnamese Social Media Text Processing.* 2023 Conference on Empirical Methods in Natural Language Processing, 2023: p. 5191–5207.

26. Tran, T.O. and P. Le Hong. *Improving sequence tagging for Vietnamese text using transformer-based neural models*. in *Proceedings of the 34th Pacific Asia conference on language, information and computation*. 2020.
27. Clark, K., et al., *Electra: Pre-training text encoders as discriminators rather than generators.* arXiv preprint arXiv:2003.10555, 2020.
28. El-Rashidy, M.A., et al., *Attention-based contextual local and global features for urgent posts classification in MOOCs discussion forums.* Ain Shams Engineering Journal, 2023: p. 102605.
29. Khodeir, N.A., *Bi-GRU urgent classification for MOOC discussion forums based on BERT.* IEEE Access, 2021. **9**: p. 58243-58255.
30. Guo, S.X., et al., *Attention-based character-word hybrid neural networks with semantic and structural information for identifying of urgent posts in MOOC discussion forums.* IEEE access, 2019. **7**: p. 120522-120532.
31. Joshy, K., R. Thakurta, and A.A. Sekh, *Future educational environment–Identification of smart campus topic trends using text mining.* International Journal of Educational Management, 2023. **37**(4): p. 884-906.
32. Nguyen, L.T., K. Van Nguyen, and N.L.-T. Nguyen, *Constructive and toxic speech detection for open-domain social media comments in vietnamese.* Advances and Trends in Artificial Intelligence. Artificial Intelligence Practices: 34th International Conference on Industrial, Engineering and Other Applications of Applied Intelligent Systems, IEA/AIE 2021, Kuala Lumpur, Malaysia, July 26–29, 2021, Proceedings, Part I 34, 2021: p. 572-583.
33. Luu, S.T., K.V. Nguyen, and N.L.-T. Nguyen, *A large-scale dataset for hate speech detection on vietnamese social media texts.* Advances and Trends in Artificial Intelligence. Artificial Intelligence Practices: 34th International Conference on Industrial, Engineering and Other Applications of Applied Intelligent Systems, IEA/AIE 2021, Kuala Lumpur, Malaysia, July 26–29, 2021, Proceedings, Part I 34, 2021: p. 415-426.
34. Truong, T.-L., H.-L. Le, and T.-P. Le-Dang, *Sentiment analysis implementing BERT-based pre-trained language model for Vietnamese.* 2020 7th NAFOSTED Conference on Information and Computer Science (NICS), 2020: p. 362-367.
35. Nguyen, H.T., et al., *VLSP shared task: sentiment analysis.* Journal of Computer Science and Cybernetics, 2018. **34**(4): p. 295-310.
36. Nguyen, K.V., et al., *UIT-VSFC: Vietnamese students' feedback corpus for sentiment analysis.* 2018 10th international conference on knowledge and systems engineering (KSE), 2018: p. 19-24.
37. Yao, Y., et al., *A survey on large language model (llm) security and privacy: The good, the bad, and the ugly.* High-Confidence Computing, 2024: p. 100211.
38. Sun, C., et al., *How to fine-tune bert for text classification?* Chinese Computational Linguistics: 18th China National Conference, CCL 2019, Kunming, China, October 18–20, 2019, Proceedings 18, 2019: p. 194-206.
39. Yu, S., J. Su, and D. Luo, *Improving bert-based text classification with auxiliary sentence and domain knowledge.* IEEE Access, 2019. **7**: p. 176600-176612.
40. Li, Q., et al., *A survey on text classification: From traditional to deep learning.* ACM Transactions on Intelligent Systems and Technology (TIST), 2022. **13**(2): p. 1-41.
41. Minaee, S., et al., *Deep learning--based text classification: a comprehensive review.* ACM computing surveys (CSUR), 2021. **54**(3): p. 1-40.
42. Dong, J., et al., *A commodity review sentiment analysis based on BERT-CNN model.* 2020 5th International conference on computer and communication systems (ICCCS), 2020: p. 143-147.
43. Kaur, K. and P. Kaur, *BERT-CNN: improving BERT for requirements classification using CNN.* Procedia Computer Science, 2023. **218**: p. 2604-2611.

44. Liu, X., et al., *Multi-task deep neural networks for natural language understanding.* Annual Meeting of the Association for Computational Linguistics, 2019.
45. Bi, Q., et al., *Mtrec: Multi-task learning over bert for news recommendation.* Findings of the Association for Computational Linguistics: ACL 2022, 2022: p. 2663-2669.
46. McCann, B., et al., *The natural language decathlon: Multitask learning as question answering.* arXiv preprint arXiv:1806.08730, 2018.
47. Jiang, H., et al., *Smart: Robust and efficient fine-tuning for pre-trained natural language models through principled regularized optimization.* 58th Annual Meeting of the Association for Computational Linguistics, 2020: p. 2177–2190.
48. Nguyen, D.Q. and A.T. Nguyen, *PhoBERT: Pre-trained language models for Vietnamese.* arXiv preprint arXiv:2003.00744, 2020.
49. Wettig, A., T. Gao, and Z. Zhong, *Should You Mask 15% in Masked Language Modeling?. arXiv.* 17th Conference of the European Chapter of the Association for Computational Linguistics, 2022: p. 2985–3000.
50. Hurst, A., et al., *Gpt-4o system card.* arXiv preprint arXiv:2410.21276, 2024.
51. Anthropic. *Introducing Claude 4*. 2025; Available from: https://www.anthropic.com/news/claude-4.
52. Phan, C.-T., et al., *ViCGCN: Graph Convolutional Network with Contextualized Language Models for Social Media Mining in Vietnamese.* arXiv preprint arXiv:2309.02902, 2023.
53. Nguyen, L.T., *Vihatet5: Enhancing hate speech detection in vietnamese with a unified text-to-text transformer model.* arXiv preprint arXiv:2405.14141, 2024.
54. Tran, Q.K., et al., *Vietnamese hate and offensive detection using PhoBERT-CNN and social media streaming data.* Neural Computing and Applications, 2023. **35**(1): p. 573-594.
55. Doan, A.L. and S.T. Luu, *Improving sentiment analysis by emotion lexicon approach on vietnamese texts.* 2022 International Conference on Asian Language Processing (IALP), 2022: p. 39-44.
56. Phạm, T.H., et al., *Ảnh hưởng của cảm nhận về rủi ro bảo mật và quyền riêng tư đến niềm tin và hành vi kiểm soát quyền riêng tư của người dùng trên mạng xã hội.* Tạp chí Kinh tế và Phát triển, 2023(314): p. 35-45.
57. Yang, A., et al., *Qwen3 technical report.* arXiv preprint arXiv:2505.09388, 2025.
58. Grattafiori, A., et al., *The llama 3 herd of models.* arXiv preprint arXiv:2407.21783, 2024.
59. Yu, T., et al., *Gradient surgery for multi-task learning.* Advances in neural information processing systems, 2020. **33**: p. 5824-5836.
60. Houlsby, N., et al., *Parameter-efficient transfer learning for NLP.* International conference on machine learning, 2019: p. 2790-2799.
61. Pfeiffer, J., et al., *Adapterfusion: Non-destructive task composition for transfer learning.* 16th Conference of the European Chapter of the Association for Computational Linguistics: Main Volume, 2021: p. 487–503.